\documentclass[10pt,twocolumn,letterpaper]{article}

\usepackage{cvpr}
\usepackage{times}
\usepackage{epsfig}
\usepackage{graphicx}
\usepackage{amsmath}
\usepackage{amssymb}
\usepackage{multirow}

\makeatletter
\newcommand{\thickhline}{\noalign {\ifnum 0=`}\fi \hrule height 1pt \futurelet \reserved@a \@xhline}
\newcommand{\tableFont}{\fontsize{7.5pt}{10pt}\selectfont}

\usepackage[breaklinks=true,bookmarks=false]{hyperref}

\cvprfinalcopy % *** Uncomment this line for the final submission

\def\cvprPaperID{****} % *** Enter the CVPR Paper ID here

\begin{document}

%%%%%%%%% TITLE
\title{Learning to Detect Instantaneous Changes with Retrospective Convolution and Static Sample Synthesis}

\author{
    Chao Chen\textsuperscript{1}, Sheng Zhang\textsuperscript{2} \thanks {Corresponding author}, Cuibing Du\textsuperscript{1}\\
    \textsuperscript{1}Department of Electronic Engineering, Tsinghua University, Beijing, China\\
    {\tt\small \{c-chen16, dcb16\}@mails.tsinghua.edu.cn}\\
    \textsuperscript{2}Advanced Sensor and Integrated System Lab\\
    Graduate School at Shenzhen, Tsinghua University, Shenzhen, China\\
    {\tt\small zhangsh@sz.tsinghua.edu.cn}
}

\maketitle
%\thispagestyle{empty}

%%%%%%%%% ABSTRACT
\begin{abstract}
    Change detection has been a challenging visual task due to the dynamic nature of real-world scenes. Good performance of existing methods depends largely on prior background images or a long-term observation. These methods, however, suffer severe degradation when they are applied to detection of instantaneously occurred changes with only a few preceding frames provided. In this paper, we exploit spatio-temporal convolutional networks to address this challenge, and propose a novel retrospective convolution, which features efficient change information extraction between the current frame and frames from historical observation. To address the problem of foreground-specific over-fitting in learning-based methods, we further propose a data augmentation method, named static sample synthesis, to guide the network to focus on learning change-cued information rather than specific spatial features of foreground. Trained end-to-end with complex scenarios, our framework proves to be accurate in detecting instantaneous changes and robust in combating diverse noises. Extensive experiments demonstrate that our proposed method significantly outperforms existing methods.
\end{abstract}

%%%%%%%%% BODY TEXT
\section{Introduction}
    Change detection plays an important role in the field of computer vision. It aims to detect saliently changing or moving regions at the pixel level, and often serves as a trigger event or a pre-processing stage in a wide variety of higher-level computer vision applications concerning video analysis.
    \begin{figure}[t]
        \centering
        \includegraphics[width=8cm]{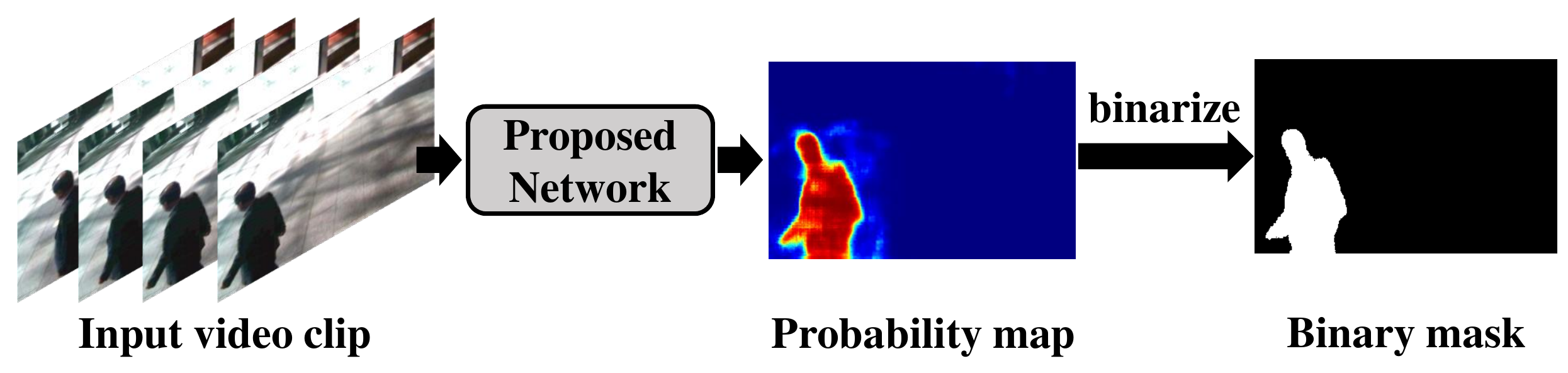}
        \caption{Illustration of our work on instantaneous change detection.}
        \label{fig:figure1}
    \end{figure}
\par
    In traditional definition of change detection,  it is assumed that there is a ¡°steady and objective¡± background scene, and disparities with it are considered as foreground of interest. But in some occasions the so-call background scene is only temporary because previously changing foreground may become stationary and turn into background for following observation. In other occasions, changing foreground may appear from the very beginning of observation and no prior background scene is available. What we are more concerned about in these cases are the instantaneously occurred changes based on the most recent observation irrespective of its past states.
\par
    Change detection on real-world scenes faces unfavorable conditions, such as illumination variation, low visibility, dynamic behavior of background, etc. Instantaneous change detection further requires these difficulties well handled based only on observation of several preceding frames.
\par
    A major category of change detection methods is background subtraction (BGS). BGS-based methods \cite{mog2, vibe, pbas, subsense} normally follow a pattern, in which a background model is built in advance and then used as the reference for foreground/background classification. However, a robust background model relies on background images or a sufficient observation on the current scenario. Another genre is based on low rank and sparse decomposition (LRS), which treats foreground and background respectively as the low-rank and sparse components of a matrix and separate them via matrix decomposition in a batch manner \cite{rpca, decolor, mogrpca}. But well-performing results require well satisfaction to their assumptions on foreground and background representations which often go unmet in real-world scenarios. Other methods \cite{opseg, ptseg} depend on quality optical flow or point trajectories and are inadequate in dealing with background dynamics.
 \par
    In this paper, we propose an end-to-end convolutional network to address the challenges faced in instantaneous change detection. Our framework takes as input a short video clip consisting of the current frame and its preceding frames, and yields per-pixel prediction of change-cued regions, as illustrated in Figure \ref{fig:figure1}. Our contributions in this paper can be summarized as follows:

    \begin{itemize}

    \item We propose a novel retrospective convolution that directly links the current frame to any preceding frame in spite of time intervals and shows effectiveness and efficiency in excavating instantaneous change information. An atrous retrospective pyramid pooling (ARPP) module is further proposed to enhance retrospective convolution with multi-scale field-of-views.

	\item To address the problem of foreground-specific over-fitting that the network might falsely respond to non-changing foreground, we propose a method named static sample synthesis that guides the network to learn change-triggered features.
	
    \item An end-to-end framework is developed to fuse change features of different scales and realize accurate per-pixel prediction.

    \item Experimental results on challenging scenarios in CDnet 2014 \cite{cdnet2014} shows obvious superiority of our framework to other approaches.

    \end{itemize}
\par
    The rest of this paper is organized as follows. Relevant work is discussed in Section \ref{section:2}. Details about our proposed methods are illustrated explicitly in Section \ref{section:3}. Data preparation and experiments are respectively described in Section \ref{section:4} and Section \ref{section:5}, followed by conclusions in Section \ref{section:6}.
\section{Related Works} \label{section:2}
    In this section we review some of the most representative works on change detection, as well as relevant convolutional networks that inspire our work.
\par
    Background subtraction based methods initialize a background model using a frame sequence from historical observation and detect change foreground of the current frame based on disparities with the background model. \cite{mog} models background on pixel values with a mixture of Gaussian models, and its improved version \cite{mog2} allows automatically selection of the proper number of components. Its variants are widely adopted in recent works \cite{ftsg, stgmm}. Alternatively, \cite{vibe} chooses to build a sample set that stores historical background values for each pixel position and use it to classify new-coming pixel values. The set is also contiguously updated by new background values so as to adapt to background dynamics. \cite{pbas} improves this method using a feedback mechanism for parameter self-adjustment. Further improvement is raised by \cite{subsense}, in which a LBSP descriptor rather than raw pixel values is used to enhance robustness against illumination variations. More discussion on background subtraction can be seen in surveys \cite{reviewbgs, surveybm, overviewbm}. Relying on prior background knowledge has been a primary obstacle restricting extensive application of BGS-based methods.
\par
    Low-rank and sparse matrix decomposition based methods offer a different methodology. As proposed by \cite{rpca}, a matrix can be decomposed into two components, a low-rank matrix and a sparse one, using Principal Component Pursuit (PCP). The method can be applied to change detection based on the assumption that background are relatively static and foreground can be sparsely represented. Further modifications and additional constrains have been developed based on this basic framework to cope with various conditions, as in  \cite{stocrpca, mogrpca, decolor, godec, blockrpca} (refer to surveys \cite{reviewrpca, reviewlrs} for detailed comparison and analysis). LRS-based methods show advantages due to no need of background modeling beforehand and hence less dependency on background images, but suffers severe degradation when real-life scenarios do not suffice to their assumptions, especially in the presence of dynamic background and slowly moving foreground.
\par
    Recently, CNN based architectures have been exploited for change detection. \cite{interactivedl, fgbgs} trains a CNN model to learn foreground representation using part of video frames, and apply inference on the rest. Despite outstanding performance, the disability of spatial CNNs in representing temporal information confines their application in specific scenes used during training. In \cite{deepbs, dbs, matchnet}, CNN models are used as the foreground/background classifier that compares a  pair of images, background image and the current frame, to find changes. They still fall into the category of background subtraction due to their dependency of prior background image.
\par
    \cite{3dbs, lstmbs} start to disengage from dependency of background images and use 3D convolution \cite{3dhuman, c3d} to learn temporal information. However, their works restrict testing scenarios to those used for training and take no consideration for foreground-specific over-fitting issue. Although spatio-temporal frameworks are used, it is still unknown whether the models have learned valid change-related features or specific spatial features of foreground as in \cite{interactivedl, fgbgs}. It is also unknown whether the well-trained models apply only to specific scenes or work just as well on scenes never before seen during training. In our wok, we propose a more efficient retrospective convolution to represent change information, and address the foreground-specific over-fitting issue with static sample synthesis method. Accurate inferences on scenes unused during training verify the generalization of our framework.
\section{Methods} \label{section:3}
    The key factors to the success of our framework in instantaneous change detection are two-fold, retrospective convolution that enables effective extraction of inter-frame change features (see Section \ref{section:3.1}) and static sample synthesis that overcomes the over-fitting of CNN-based framework to foreground-specific spatial features (see Section \ref{section:3.2}). We also utilize a deep spatial convolutional network shared among frames to learn multi-scale change-aware spatial features of foreground (see Section \ref{section:3.3}), and adopt an encoder-decoder structure to fuse change features of different scales and realize dense prediction at the input resolution (see Section \ref{section:3.4}).
    \subsection{Retrospective convolutions for extracting change-cued temporal information} \label{section:3.1}
        How to extract change-targeted temporal information from a short observation is a primary challenge in instantaneous change detection. In video understanding tasks such as action recognition, 3D convolutions are used to extract inter-frame information and have successfully shown their power in spatio-temporal feature representation. Our preliminary experiments using 3D convolutions also testify the feasibility of extending 3D convolutions to instantaneous change detection (see experiments in Section \ref{section:5.1}).
        \begin{figure*}[t]
            \centering
            \includegraphics[width=14cm]{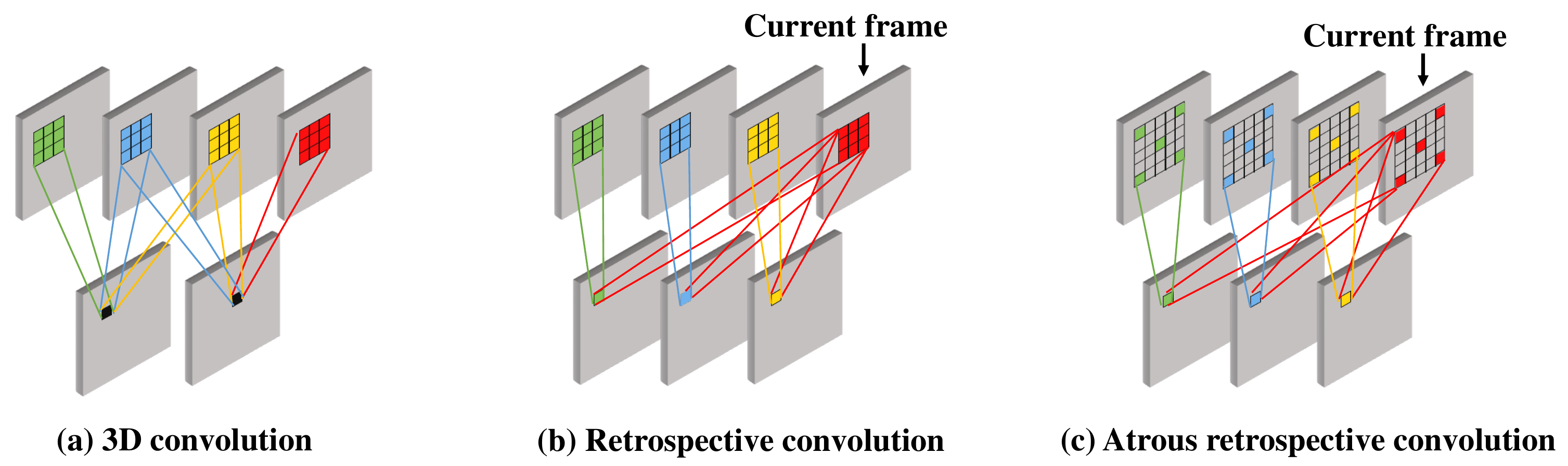}
            \caption{Comparison of 3D convolution, retrospective convolution and atrous retrospective convolution. (a) A 3D convolution kernel of size $3 \times 3 \times 3$ works on three consecutive frames, and a frame can not be linked directly to another one with more than 2-frame interval. (b) A retrospective convolution kernel of spatial size $3 \times 3$ relate the current frame to each of all preceding frames (c) An atrous retrospective convolution kernel with dilation = 2 expands the FoV from $3 \times 3$ to $5 \times 5$.}
            \label{fig:figure2}
        \end{figure*}
    \par
        But there are fundamental distinctions between change detection and action recognition-like tasks. In change detection, we look for changing regions in the current frame and concern with the underlying disparity and consistency between the current frame and historical observation. 3D convolutions, however, work on locally contiguous frames and it would have to use cascaded layers or larger kernels to bridge frames with large temporal intervals. As shown in Figure \ref{fig:figure2}(a), it needs at least two cascaded layers of $3 \times 3 \times 3$ 3D convolution to link the current frame to the farthest historical frame in the case of 4-frame video clip. Besides, 3D convolutions that do not involve the current frame might produce redundant temporal information and cause distraction and interference during the learning process.
    \par
        \textbf{Notations.} We follow the convention in \cite{c3d} of expressing kernel size in $l \times h \times w$, where $l$ for temporal length of frames and $h \times w$ for spatial resolution. For a frame sequence of length $L$, we refer to the first $L-1$ frames ($l = 0, \dots , L - 2$) as historical observation and the last one ($l = L - 1$) as the current frame.
    \par
        \textbf{Retrospective convolutions.} We propose instead the use of retrospective convolutions that break away from the temporal limitation of 3D convolutions and allow direct and dense connectivity between the current frame and all the rest of the video clip (as illustrated in Figure \ref{fig:figure2}(b)). Formally, the value at location $(l,i,j)$ on the $d$th resulting feature map using a retrospective convolution and an activation function can be given by:
        \begin{equation}
        \begin{split}
            x_{d,l,i,j}^{(out)} =& f_{acti}(\sum_{c=0}^{C-1}\sum_{h=0}^{H-1}\sum_{w=0}^{W-1} w_{c,0,h,w} x_{c,l,i+h,j+w}^{(in)} \\
            +& \sum_{c=0}^{C-1}\sum_{h=0}^{H-1}\sum_{w=0}^{W-1} w_{c,1,h,w} x_{c.L-1,i+h,j+w}^{(in)})
        \end{split}
        \label{formula1}
        \end{equation}
        where $w$ is the weight of retrospective convolution kernel with size $2 \times H \times W$.
    \par
        As seen from Formula \ref{formula1}, a retrospective convolutional kernel can be considered as the combination of two spatial convolutional kernels. The one half with $l = 1$ is assigned to the current frame and the other half with $l = 0$ to a historical frame. The responses at $l = l_0 (l_0 = 0, \dots , L - 2)$ are the inter-frame features of frames at $l = l_0$ and $l = L-1$. In this manner, the current frame is enabled to fully associate with each historical frame despite their temporal intervals within a single retrospective convolutional layer.
    \par
\begin{figure}[t]
            \centering
            \includegraphics[width=8cm]{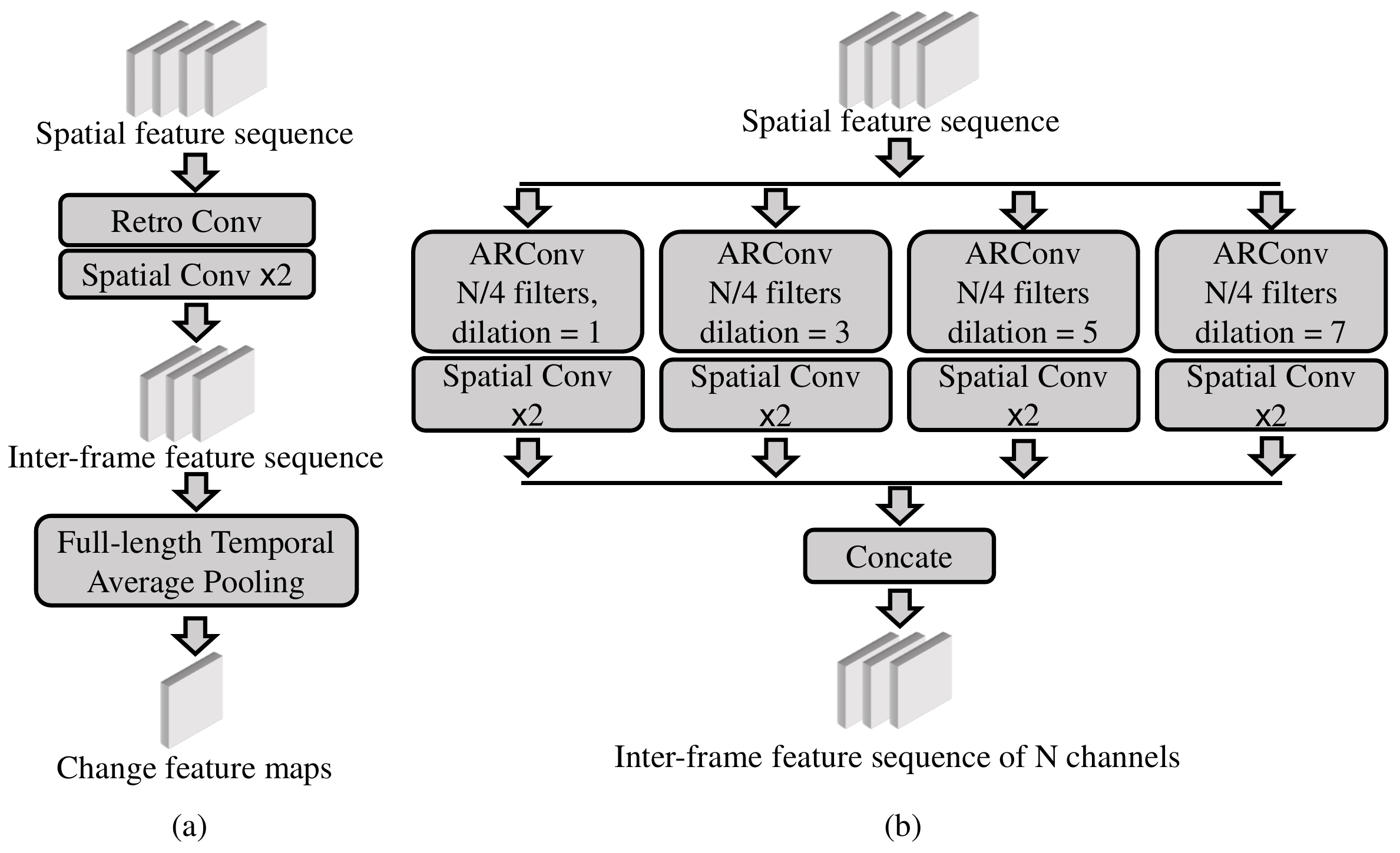}
            \caption{(a) A retrospective convolution module for change feature map extraction, which consists of a layer of retrospective convolution followed by two layers of spatial convolution and a full-length temporal average pooling. (b) An example of ARPP. In this structure, there are four parallel groups of ARConv with \{1, 3, 5, 7\} dilation respectively.}
            \label{fig:figure3}
        \end{figure}

        \textbf{Extraction of change feature map.} Figure \ref{fig:figure3}(a) illustrates the structure used in our work to extract change feature map. The input spatial feature sequence is firstly processed with a layer of retrospective convolutions, and then two spatial convolutional layers for enhanced representation of inter-frame features. A full-length temporal average pooling is finally used to summarize all inter-frame features, which simply computes the mean values on the time axis:
        \begin{equation}
            x_{c,h,w}^{(out)} = \frac{1}{L-1} \sum_{l=0}^{L-2} x_{c,l,h,w}^{(in)}
        \end{equation}
    \par
        In a sense, the output change feature map stands for the average performance of all retrospectively observed inter-frame features.
    \par
        The proposed structure is temporally scalable. A trained network using this structure can be directly applied to a video clip of any length with no need of any network reconfiguration or finetuning.
    \par
        \textbf{Atrous retrospective convolution and atrous retrospective pyramid pooling.} Inspired by atrous convolution and ASPP structure proposed in \cite{deeplabv2}, we also take into account multi-scale field-of-views and borrow ``atrous'' to extend retrospective convolution. Atrous retrospective convolution (ARConv), as illustrated in Figure \ref{fig:figure2}(c), employs dilated spatial kernels and enlarge the spatial size from $k$ to $k + (k - 1)(dilation - 1)$.
    \par
        ARConvs with different dilations can also be deployed in parallel to obtain multi-scale spatial FoVs (we name it ARPP, short for ``Atrous Retrospective Pyramid Pooling''). Figure \ref{fig:figure3}(b) shows an ARPP module with 4 branches, each branch using a different dilation. Compared to a retrospective convolution module of $N$ filters, an $m$-branch ARPP module uses $N/m$ filters in each branch and acquires multiple FoVs with less parameters (the reduced parameters come from the spatial convolutions in the module).
    \subsection{Static sample synthesis against foreground-specific over-fitting} \label{section:3.2}
        A change detection network is learning both spatial and temporal representations during training and chances are that a trained network may respond positively not only to changing foreground but also to static foreground with specific spatial features (i.e. the foreground-specific over-fitting phenomenon).
        \begin{figure}[t]
            \centering
            \includegraphics[width=8cm]{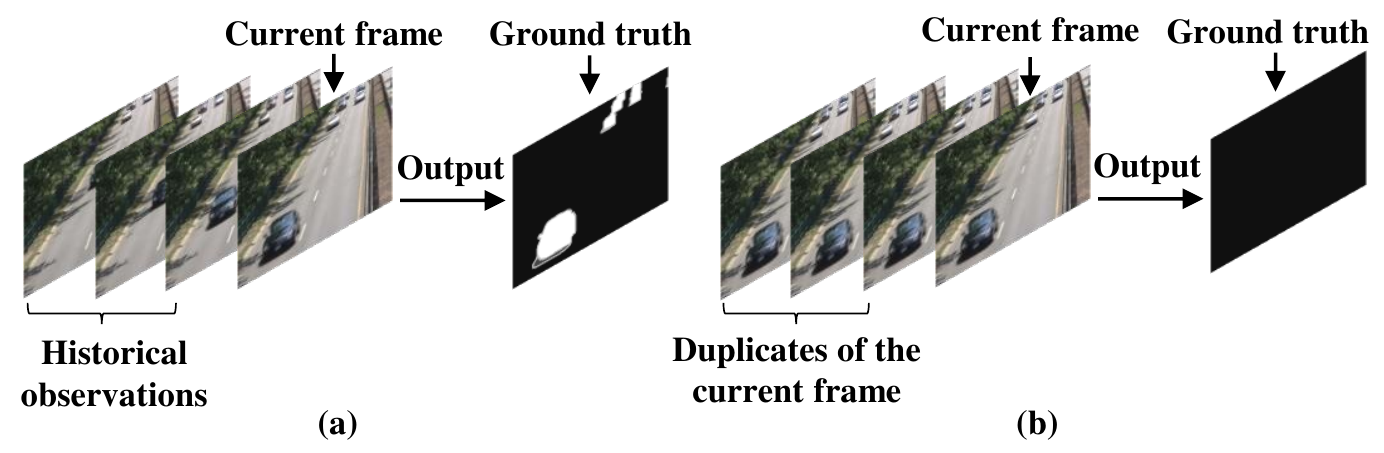}
            \caption{A native sample (a) and its corresponding synthesized static sample (b).}
            \label{fig:figure4}
        \end{figure}
        \begin{figure}[t]
            \centering
            \includegraphics[width=8cm]{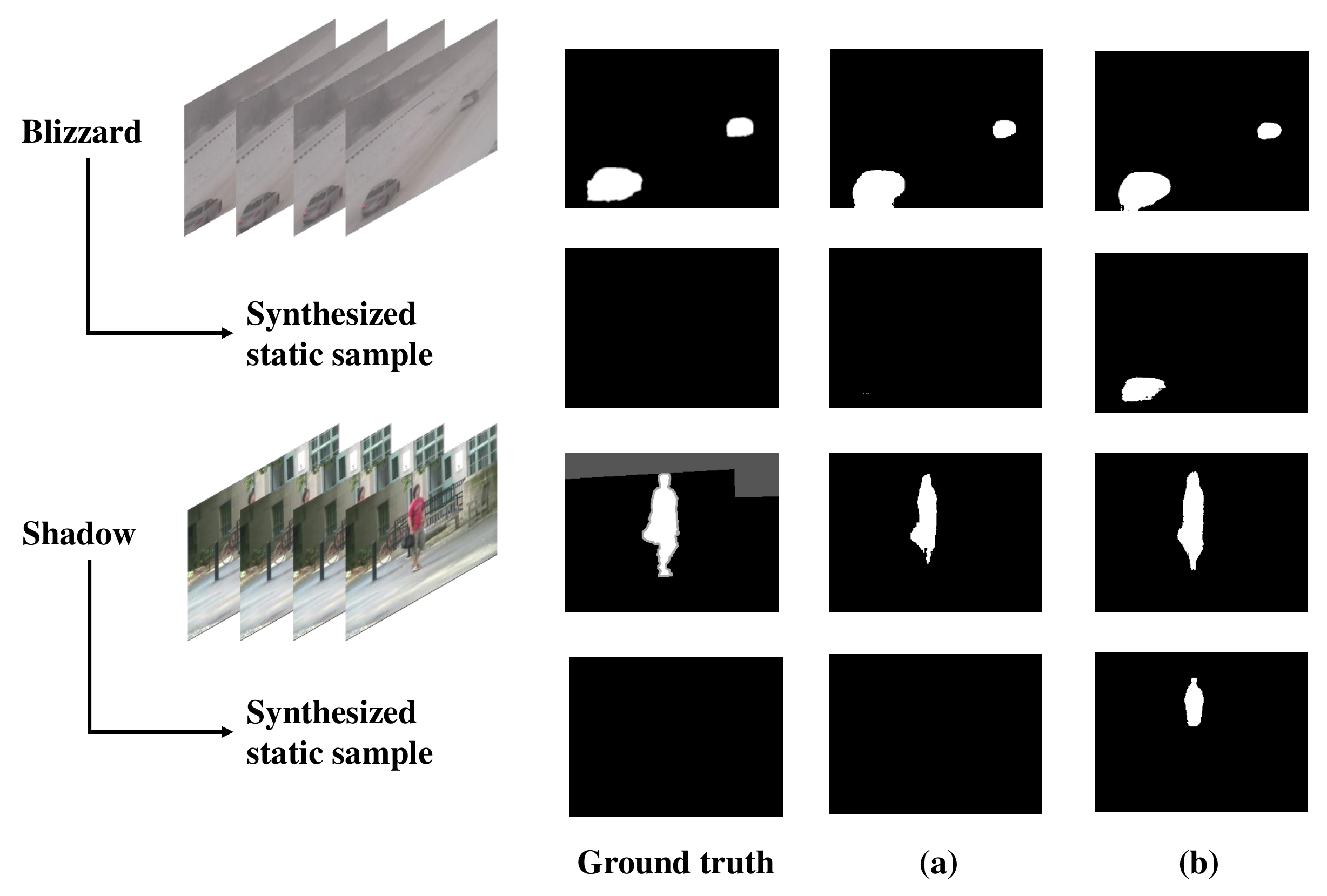}
            \caption{The foreground-specific over-fitting phenomenon and the effect of training with synthesized static samples. (a) Results of the network trained with extra synthesized static samples. (b) Results of the network trained only with native samples.}
            \label{fig:figure5}
        \end{figure}
    \par
        We advocate the use of static sample synthesis method to address the problem. For each native input video clip with changing foreground, a completely static sample is synthesized in the way that all historical frames are simply duplicated from the current frame (see Figure \ref{fig:figure4}). The method works as a guidance for training and strongly suppresses false responses to non-changing foreground. Figure \ref{fig:figure5} visualizes the difference between models trained with or without synthesized static samples.
    \subsection{A shared spatial convolutional network for change-aware spatial feature learning} \label{section:3.3}
        We use a spatial convolutional network for richer spatial feature representation in advance of change feature extraction. The spatial convolutions are implemented in the manner of 3D convolutions so as to be shared among all frames. Specifically, a $1 \times k \times k$ 3D convolution is used to realize a de facto $k \times k$ spatial convolution. Due to pure inner-frame operations in this part, temporal length of features stays the same as input in all layers.
    \par
        Different layers in a deep spatial convolutional network yield spatial features with receptive fields of a wide range of scales, and can provide diverse perspectives for the following change feature extraction. Spatial features from lower-levels have high resolutions and detailed descriptions, which enable sensitive perception of small objects and slight changes. Higher-level features, on the other hand, go with coarser resolutions but stronger semantic representation, which helps to observe object-level changes.
        \begin{figure*}[t]
            \centering
            \includegraphics[width=15cm]{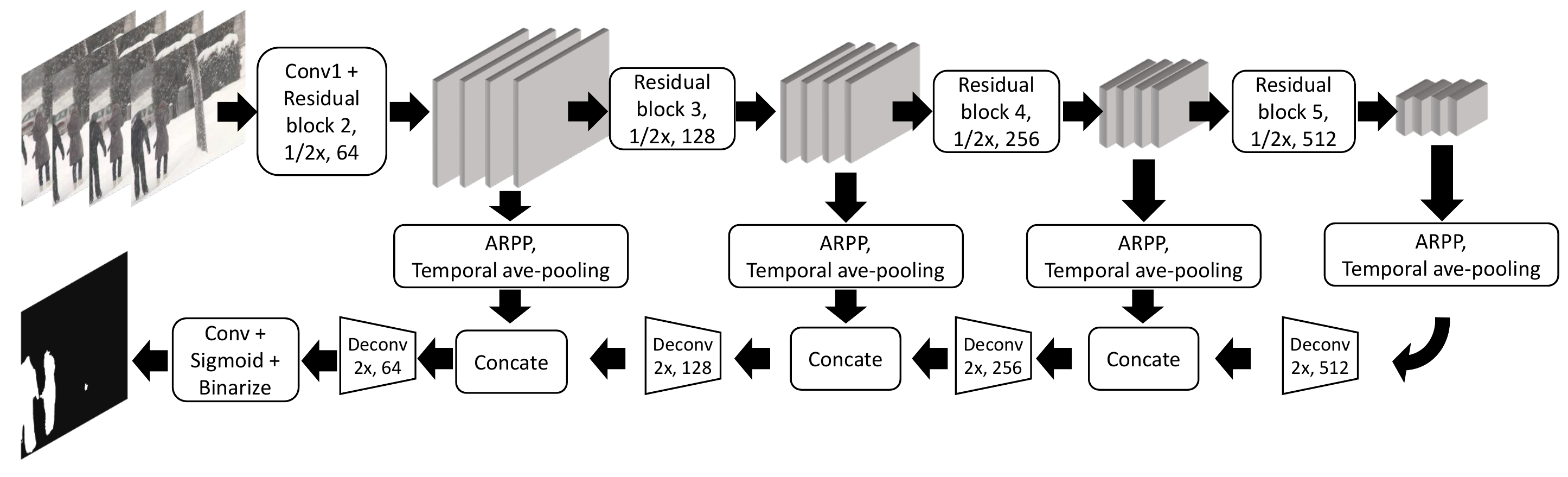}
            \caption{Architecture of our proposed network based on ResNet-18, ARPP and multi-level encoder-decoder modules}
            \label{fig:figure6}
        \end{figure*}
    \subsection{An encoder-decoder structure for multi-scale change information fusion} \label{section:3.4}
        Change detection has been facing a pair of contradicting problems. On one side, lower-level features are good at sensing local changes, especially marginal area of objects, but are easily affected by background noises and are also prone to holed or fragmental results. On the other side, higher-level features provide object-level or frame-level representation but become less sensitive to partial movement of non-rigid objects and slowly moving objects. With consideration to both situations, we use an encoder-decoder structure to fuse change features at multiple levels in a top-down manner. In each decoder module, higher-level change feature maps is up-scaled with a $2 \times 2$ deconvolution with stride 2 and then concatenated with features at the lower-level.  With multi-level decoding, change features is finally obtained at the input resolution and used for per-pixel prediction. The overall architecture is illustrated in Figure \ref{fig:figure6}.
\section{Dataset} \label{section:4}
    \subsection{CDnet 2014} \label{section:4.1}
        We evaluate our method on CDnet 2014 dataset \cite{cdnet2014}, a large real-scene video dataset designated for change detection task. It contains 53 videos with pixel-wise annotated frames, including 11 categories that cover various challenging scenarios (e.g., bad weather, night, dynamic background) in practical applications.
\par
        Note that in conventional definition of change detection as in CDnet 2014, changes relative to scenes at the beginning of a video are reckoned as foreground of interest and will remains as foreground throughout the video even they become completely stationary after appearance (as samples in the category ¡°Intermittent Object Motion¡±). In instantaneous change detection, however, we only look at the most recent several frames and detect changes according to them. Considering this divergence, we abandon samples with no recent changes in our experiments to avoid ambiguity.
\par
        Each frame in the videos, along with its several preceding frames referred as historical observation, constitutes an input video clip sample (as shown in Figure \ref{fig:figure2}(a)). Videos selected from CDnet 2014 for training and testing are listed in Table \ref{tab:1}.
        \begin{table}[t]
        \centering
        \tableFont
        \begin{tabular}{p{2.5cm}|p{1.8cm}|p{2.2cm}}
            \thickhline
            Categories  & Training      & Testing       \\ \thickhline

            ~           & blizzard      & ~             \\
            badWeather  & skating       & snowFall      \\
            ~           & wetSnow       & ~             \\ \hline

            ~           & highway       & ~              \\
            baseline    & office        & PETS2006      \\
            ~           & pedestrains   & ~             \\ \hline

            \multirow{4}*{dynamicBackground} & boats & ~ \\
            ~           & fall          & fountain02     \\
            ~           & fountain01    & canoe          \\
            ~           & overpass      & ~              \\ \hline

            \multirow{4}*{nightVideos}   & bridgeEntry & ~         \\
            ~           & busyBoulvard   & streetCornerAtNight     \\
            ~           & fluidHighway   & winterStreet            \\
            ~           & tramStation    & ~                       \\ \hline

            \multirow{4}*{shadow}        & backdoor & ~     \\
            ~           & busStation     & peopleInShade    \\
            ~           & copyMachine    & bungalows        \\
            ~           & cubicle        & ~                \\ \thickhline
        \end{tabular}
        \caption{Vidoe categories in CDnet 2014 used for training and testing.}
        \label{tab:1}
        \end{table}
    \subsection{Training data pre-processing} \label{section:4.2}
        Video clips with various spatial scales or temporal length are all feasible for training. As a trade-off between efficiency and accuracy, we choose to use input of resolution $128 \times 160$ and length 4.
\par
        Commonly seen data augmentation strategies are utilized, such as mean-subtraction, randomly horizontal and vertical flipping, contrast and brightness jittering, noise addition, etc. More strategies, multi-scale cropping, class balancing, scenario balancing, and temporal sampling interval jittering, are specially designed to pre-process the training dataset.
\par
        \textbf{Multi-scale cropping.} We firstly resize all training video clips to multiple resolutions, specifically $\{1, 1/2, 1/4 \}$ of resolution $640 \times 512$, and then crop $160 \times 128$ clips from them with stride $80 \times 64$ and without padding.
\par
        \textbf{Foreground/background class balancing.} Note that samples selected and cropped from CDnet 2014 dataset are mostly background and suffers severe class imbalance. To relieve the situation, we select samples with positive label occupation ratio lying between 5\% and 60\%, because we need to abandon samples with barely any foreground pixels and meanwhile prevent those that are overly occupied by foreground objects. Meanwhile, we use a weighted cross entropy loss \cite{segnet} given by:
        \begin{equation}
            Loss = \alpha y \log \hat{y} +(1-y) \log (1-\hat{y})
        \end{equation}
\par
        In our experiments, we empirically set the weight $\alpha = 4.0$.
\par
        \textbf{Scenario balancing.} Number of valid samples (i.e. with proper positive label occupation ratio as mentioned above) varies greatly in different video scenarios. We equally pick samples from all scenarios during training.
\par
        \textbf{Temporal sampling interval jittering.} We randomly alter temporal sampling interval (ranging from 2 to 8) when picking preceding frames. This strategy effectively adapt the network to changes of varying extents.
        \begin{figure*}[t]
            \centering
            \includegraphics[width=17cm]{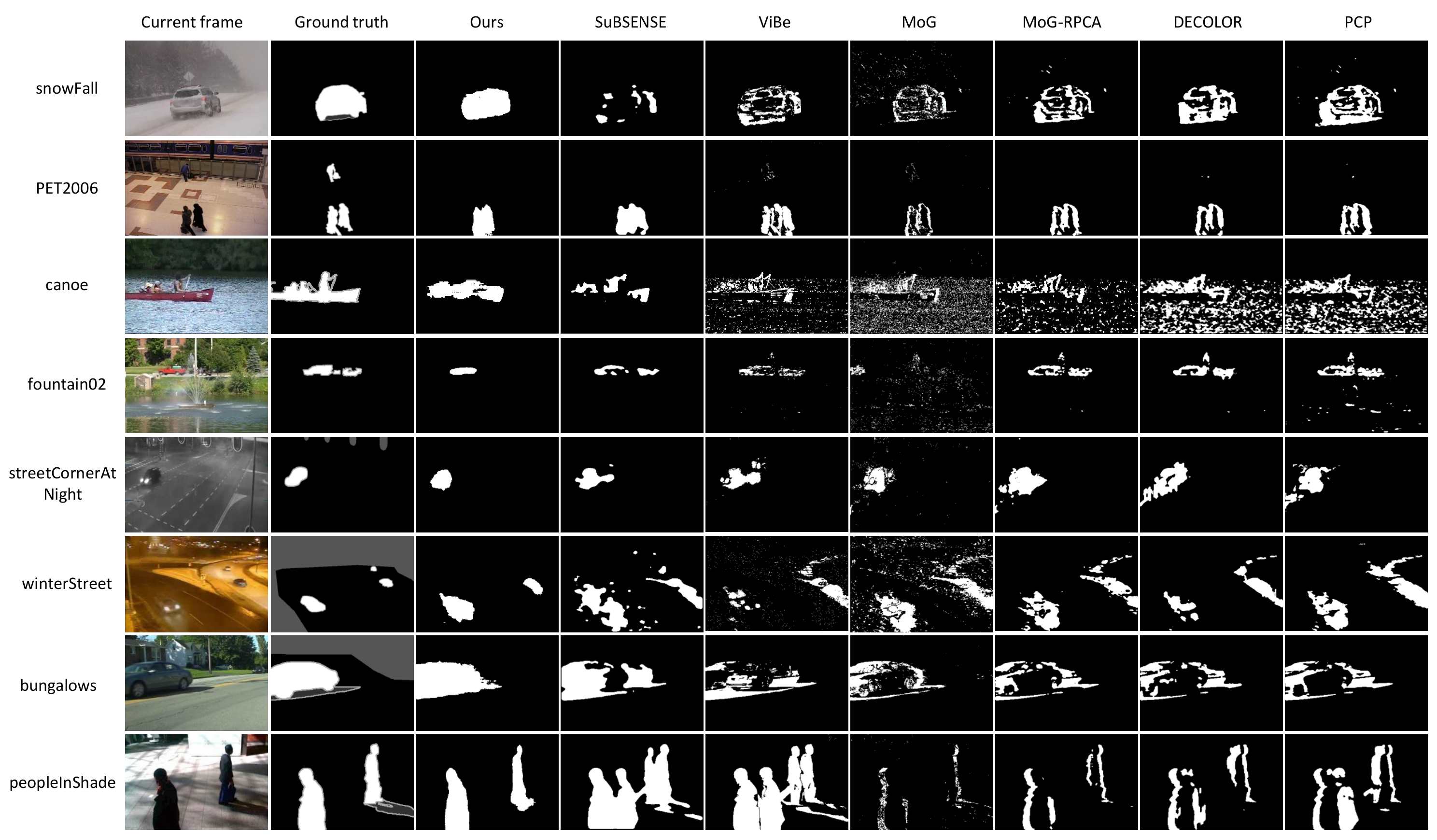}
            \caption{Qualitative results on testing video clips from different senarios in CDnet 2014.}
            \label{fig:figure7}
        \end{figure*}
\section{Experiments} \label{section:5}
    Instantaneous change detection performed by all involved methods is restricted to historical observation of no more than 6 preceding frames (corresponding to $1/4$ seconds for a 24 fps frame rate). Resolutions of different videos in CDnet 2014 vary greatly, so we choose to evaluate all testing samples at the uniform resolution of $320 \times 256$ for ease of comparison. Precision, recall and primarily F-measure (as defined in \cite{cdnet2014}) are used as metrics to quantitatively evaluate performance.
    \subsection{Evaluations of the proposed framework} \label{section:5.1}
        \textbf{Training.} Our network is trained using mini-batch SGD with a batch size of 12. Initial learning rate is set to 1e-6, and is multiplied by 0.1 every 20k iterations. Momentum is set to 0.9, and weight decay to 0.0005. Optimization process is stopped at 80k iterations where the loss normally stays steady. Models based on ResNet-18 (or part of it) are initialized with pre-trained ResNet-18 on ImageNet \cite{imagenet}. The proposed methods are implemented based on Caffe framework \cite{caffe} with our modifications.
    \par
        \textbf{Inference.} We apply multi-scale inference by separately predicting each video clip at scales \{1, 1/2\} of $320 \times 256$, and taking the mean value of probability at each position. For better computation efficiency yet without loss of accuracy, we construct testing clips for our network using 3 historical frames at temporal sampling rate 2 instead of using all 6 frames.
    \par
        \textbf{Retrospective convolutions vs. 3D convolutions} We compare the performance of frameworks using retrospective convolution and 3D convolution respectively. The 3D convolution based networks use two cascaded layers of $3 \times 3 \times 3$ 3D convolution and a full-length temporal average pooling instead of the retrospective convolution module as in Figure \ref{fig:figure3}(a). As shown in Table \ref{tab:2}, retrospective convolution brings extra 6.8\% improvement on a simple network and 3.6\% improvement on ResNet-18 \cite{resnet} based network over 3D convolutions.
        \begin{table}[t]
        \centering
        \tableFont
        \begin{tabular}{p{3cm}p{1cm}p{1cm}p{1cm}}
            \thickhline
            Methods  & Pre      & Rec   & F-M       \\ \thickhline

            Simple + 3D conv        & 0.596 & 0.588 & 0.535 \\
            Simple + Retro conv     & 0.552 & 0.766 & 0.602 \\
            ResNet-18 + 3D conv     & 0.620 & 0.714 & 0.609 \\
            ResNet-18 + Retro conv  & 0.679 & 0.708 & 0.645 \\  \thickhline
        \end{tabular}
        \caption{Comparison the influence of 3D convolutions and Retrospective convolutions. ¡®Simple¡¯ indicates the spatial convolutional network is a simple ConvNet with 3 layers of 3$\times$3 spatial convolutions with max-pooling and ReLU in between.}
        \label{tab:2}
        \end{table}
\par
        \textbf{Shared spatial convolutional network.} We report experiments with spatial convolutional networks with different depths in Table \ref{tab:3}. All networks in this part use retrospective convolution modules (Figure \ref{fig:figure3}(a)) for change feature extraction. More levels of encoder-decoder modules are utilized with depth. For example, no encoder-decoder module is needed for raw RGB input and four modules are used for a full ResNet-18 based network (as in Figure \ref{fig:figure6}).
\par
        In Table \ref{tab:3}, performance improves increasingly with more cascaded layers, and reaches a final F-measure of 64.5\%. Marginal gain of performance from depth decreases when the network goes deeper.
        \begin{table}[t]
            \centering
            \tableFont
            \begin{tabular}{p{4cm}p{0.7cm}p{0.7cm}p{0.7cm}}
                \thickhline
                Shared spatial convolutional network  & Pre      & Rec   & F-M       \\ \thickhline

                Raw RGB input                       & 0.499 &  0.543 &  0.482 \\
                ResNet-18 (first 5 conv layers)     & 0.540 &  0.739 &  0.589 \\
                ResNet-18 (first 9 conv layers)     & 0.639 &  0.719 &  0.633 \\
                ResNet-18 (first 13 conv layers)    & 0.693 &  0.683 &  0.645 \\
                ResNet-18 (all 17 conv layers)      & 0.679 &  0.708 &  0.645 \\ \thickhline
            \end{tabular}
            \caption{Performance of networks with various depths.}
            \label{tab:3}
        \end{table}
\par
        \textbf{ARPP modules.} We have experimented with the use of ARPP modules with different structures. Table \ref{tab:3} shows that the use of ARPP \{1,3\} reaches F-measure of 65.6\%, and gains 1.1\% performance promotion compared to naive retrospective convolution. With increased branches, ARPP \{1,3,5,7\} and \{1,2,3,4,5,6,7,8\} maintain the performance with further reduced parameters.
        \begin{table}[t]
            \centering
            \tableFont
            \begin{tabular}{p{4cm}p{0.7cm}p{0.7cm}p{0.7cm}}
                \thickhline
                Framework Variations & Pre      & Rec   & F-M       \\ \thickhline

                ResNet-18 + Retro conv              & 0.679 & 0.708 & 0.645 \\
                ResNet-18 + ARPP \{1,3\}              & 0.670 & 0.720 & 0.656 \\
                ResNet-18 + ARPP \{1,3,5,7\}          & 0.689 & 0.688 & 0.655 \\
                ResNet-18 + ARPP \{1,2,3,4,5,6,7,8\}  & 0.657 & 0.732 & 0.656 \\ \thickhline
            \end{tabular}
            \caption{Performance of ARPP modules. ARPP \{1,3\} means the use of an ARPP module with two branches of dilation = 1 and 3 respectively.}
            \label{tab:4}
        \end{table}
\par
        \textbf{Running time.} Our model based on ResNet-18 + ARPP \{1,3,5,7\} inferences at the speed of 77ms per 4-frame $320 \times 256$-resolution video clip on a single NVIDIA TITAN X (PASCAL) GPU. A simpler model using first 9 layers of ResNet-18 runs at 55 ms.
    \subsection{Comparison to existing methods} \label{section:5.2}
        We compare our method (a ResNet-18 + ARPP \{1,3,5,7\} framewok is used) with some of the most representative methods for change detection, include BGS-based methods SuBSENSE \cite{subsense}, Vibe \cite{vibe}, MoG \cite{mog2} and LRS-based methods MoG-RPGA \cite{mogrpca}, DECOLOR \cite{decolor}, PCP \cite{rpca}. We use implementation of these methods in BGSLibrary \cite{bgslibrary} and LRSLibrary \cite{lrslibrary}, where the source codes released by their authors are kindly collected and integrated for easy use.
    \par
        To demonstrate the performance on instantaneous change detection, BGS-based methods build background models on the preceding 6 frames and use it on the current frame for change detection (see result in Table \ref{tab:5}). LRS-based methods takes in a clip of 7 frames (the current frame and 6 preceding frames) as a whole matrix for later low-rank and sparse decomposition. Other parameters of each algorithm are all set as default. Considering that LSR-based methods are based on the existence of changing foreground, we compare our method with LRS-based methods on samples with at least 1\% foreground (see result in Table \ref{tab:6}).
        \begin{table}[t]
            \centering
            \tableFont
            \begin{tabular}{c|c|c|c|c}
                \thickhline
                Videos  & SuBSENSE    & ViBe      & MoG   &   Ours   \\ \thickhline
                snowFall        & 0.186 & 0.231 & 0.155 & \textbf{0.727} \\ \hline
                PETS2006        & \textbf{0.615} & 0.428 & 0.212 & 0.569 \\ \hline
                canoe           & 0.387 & 0.230 & 0.151 & \textbf{0.712} \\ \hline
                fountain02      & \textbf{0.612} & 0.283 & 0.331 & 0.560 \\ \hline
                streetCornerAtNight& 0.504 & 0.363 & 0.350 & \textbf{0.641} \\ \hline
                winterStreet     & 0.310 & 0.328 & 0.317 & \textbf{0.518} \\ \hline
                peopleInShadea   & 0.644 & 0.506 & 0.422 & \textbf{0.748} \\ \hline
                bungalows        & 0.572 & 0.367 & 0.453 & \textbf{0.768} \\ \hline
                Average          & 0.479 & 0.342 & 0.299 & \textbf{0.655} \\ \thickhline
            \end{tabular}
            \caption{F-Measure performance compared to BGS-based methods.}
            \label{tab:5}
        \end{table}

        \begin{table}[t]
            \centering
            \tableFont
            \begin{tabular}{c|c|c|c|c}
                \thickhline
                Videos  & MoG-RPCA    & DECOLOR      & PCP   &   Ours   \\ \thickhline
                snowFall        & 0.468 & 0.505 & 0.440 & \textbf{0.758} \\ \hline
                PETS2006        & 0.505 & 0.445 & 0.457 & \textbf{0.595} \\ \hline
                canoe           & 0.187 & 0.149 & 0.147 & \textbf{0.712} \\ \hline
                fountain02      & 0.573 & \textbf{0.583} & 0.478 & 0.561 \\ \hline
                streetCornerAtNight& 0.507 & 0.461 & 0.446 & \textbf{0.697} \\ \hline
                winterStreet     & 0.375 & 0.452 & 0.323 & \textbf{0.525} \\ \hline
                peopleInShade   & 0.379 & 0.409 & 0.454 & \textbf{0.774} \\ \hline
                bungalows        & 0.379 & 0.348 & 0.358 & \textbf{0.781} \\ \hline
                Average          & 0.422 & 0.419 & 0.388 & \textbf{0.675} \\ \thickhline
            \end{tabular}
            \caption{F-Measure performance compared to LRS-based methods (working on testing samples with at least 1\% foreground).}
            \label{tab:6}
        \end{table}
\par
        Experimental results show that existing methods work inferiorly under limited observation, despite their good performance in conventional change detection context. The main adverse factors come from slow motion and dynamic background. Figure \ref{fig:figure7} shows visual results of the compared methods.
\par
        Slow motion (as in ``snowfall'' and ``bungalows'') indicates large over-lapping of foreground among frames. The over-lapped part features insignificant change over the short-time span and often leads the existing methods to produce edge-like or holed results. Fusing low- and high-level features, our method observes changes from enlarged perspectives, which helps to yield results of complete objects. Background dynamics (as in ``canoe'') is a major source of false alarms and is easily mixed with foreground changes without long-term analysis. Our framework learns to effectively block out the interference from dynamic behaviors of background. Besides, existing methods produce obvious ghost artifacts (as in ``peopleInShade'') due to simple inter-frame comparison, but our framework considers both spatial representation and temporal relationship and effectively handles the problem.
\par
        Despite outstanding performance, our method still shows limitation in accurate segmentation of object boundaries and change detection of small foreground. The fusion of high-level features enables better observation of object-level changes but at the cost of pixel-level details.
\par
        In general, our method achieves best F-Measure performance in most testing scenarios, and outperforms the existing methods with significant average F-Measure improvement.
\par
\begin{figure}[t]
            \centering
            \includegraphics[width=6cm]{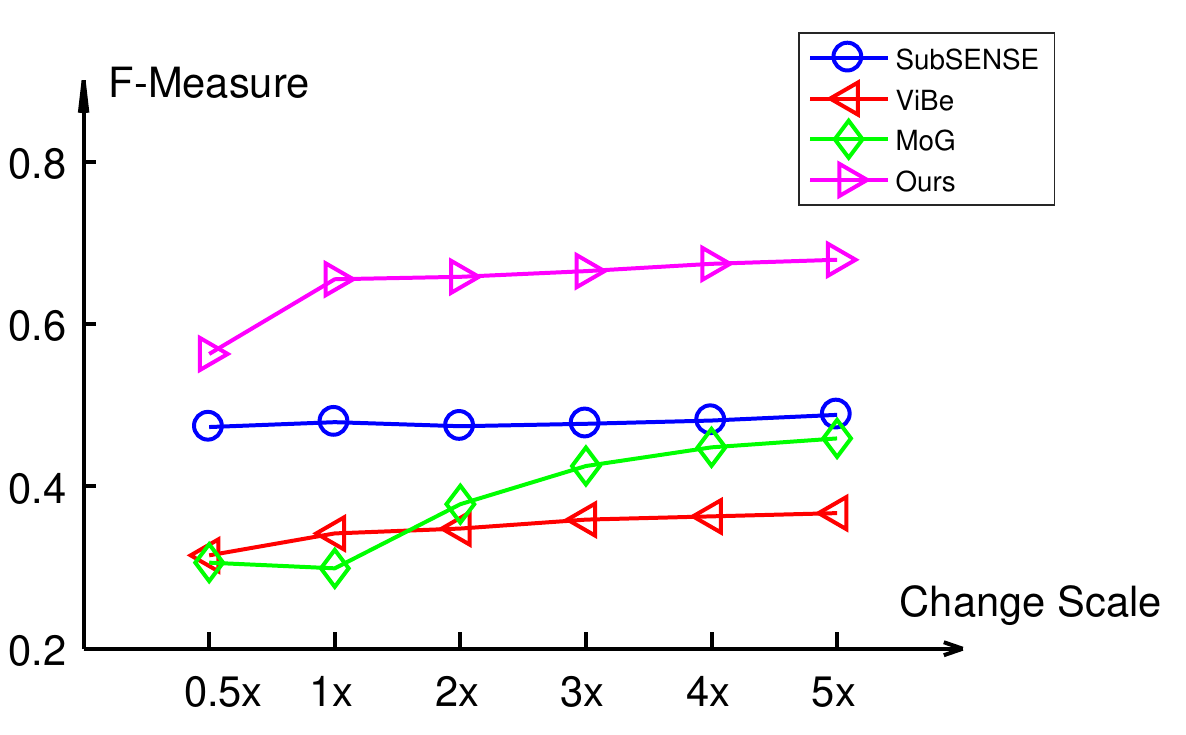}
            \caption{Performance with different change scales (compared to BGS methods).}
            \label{fig:figure8}
        \end{figure}
\begin{figure}[t]
            \centering
            \includegraphics[width=6cm]{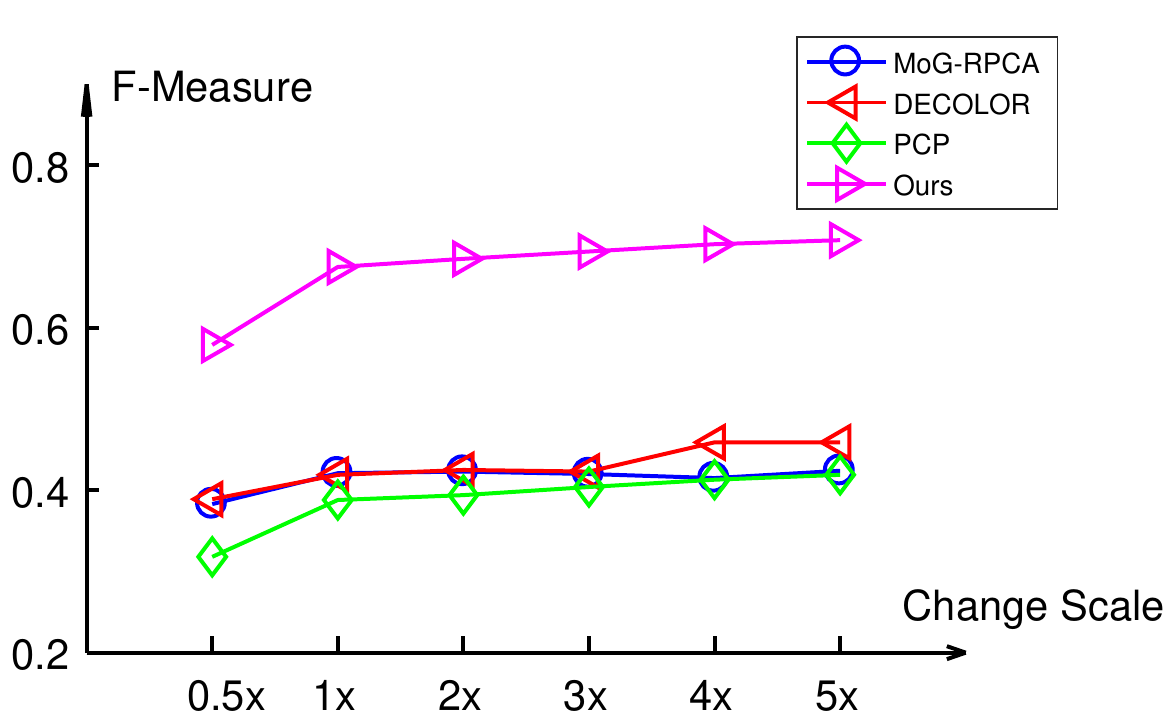}
            \caption{Performance with different change scales (compared to LRS methods).}
            \label{fig:figure9}
        \end{figure}

        \textbf{Performance on wilder changes.} To evaluate performance on wilder changes, we simulate different extents of changes by varying the temporal sampling interval for historical frames. For example, we pick the 6 preceding frames with temporal sampling interval being 2 to simulate 2x scale-up of foreground changes. Exceptionally, we use 0.5x to indicate the use of only 3 preceding frames and a scaled-down change. Figure \ref{fig:figure8}, \ref{fig:figure9} show consistent superiority of our method on a wide range of change scales.
\section{Conclusions} \label{section:6}
    In instantaneous change detection, change-cued foreground is detected based on the observation of only a few preceding frames. In this paper, we use an end-to-end framework to address the challenges of this task and propose a novel retrospective convolution that enables efficient change feature extraction. A static sample synthesis method for training data augmentation is further proposed to avoid foreground-specific over-fitting. The framework also employs shared spatial convolutions and a multi-level encoder-decoder structure to combine change features of different scales. Learning from challenging scenarios, our proposed framework show effectiveness in sensing instantaneously occurred changes and robustness against background dynamics. Experimental results demonstrate that our method significantly advances the state-of-the-art methods.
{\small
\bibliographystyle{ieee}

}

\end{document}